\documentclass[runningheads]{llncs}
\usepackage[T1]{fontenc}
\usepackage{graphicx}
\usepackage{amssymb}
\usepackage{amsmath}
\usepackage{booktabs}
\usepackage{graphicx}
\usepackage{algorithm,algorithmic}

\usepackage[nolist]{acronym}
\usepackage{color, colortbl}
\usepackage{paralist}
\usepackage{hyperref}
\usepackage{xurl}
\usepackage{xspace}
\usepackage{enumitem}

\usepackage[skins]{tcolorbox}
\usepackage{xcolor}

\tcbset{
  my box/main style/.style={},
  my box/title style/.style={},
  my box/main/.style={/tcb/my box/main style/.style={#1}},
  my box/title/.style={/tcb/my box/title style/.style={#1}},
  my box/append main/.style={/tcb/my box/main style/.append style={#1}},
  my box/append title/.style={/tcb/my box/title style/.append style={#1}},
  my box/.style={
    my box/.cd, #1,
    /tcb/.cd,
    enhanced,
    my box/main style,
    attach boxed title to top left={xshift=0.2cm, yshift=-0.2cm},
    boxed title style={
      top=1pt,
      bottom=1pt,
      my box/title style,
    },
  },
}

\newtcolorbox{finding}[2][]{
  my box={
    main={
        top=6pt,
        bottom=3pt,
        colframe=black, 
        colback=white
    },
    title={
        colframe=black, 
        colback=gray
    },
  },
  title=#2,
  #1,
}

\def\BibTeX{{\rm B\kern-.05em{\sc i\kern-.025em b}\kern-.08em
    T\kern-.1667em\lower.7ex\hbox{E}\kern-.125emX}}

\newcommand{\quotes}[1]{``#1''}
\newcommand{\sym}[1]{\textsuperscript{#1}}

\begin{document}

%%% Use this command to specify your submission number.
%%% In doubleblind mode, it will be printed on the first page.

\title{In-Context Symbolic Regression for Robustness-Improved Kolmogorov-Arnold Networks}
\titlerunning{In-Context Symbolic Regression for Robustness-Improved KANs}

\author{Francesco Sovrano\inst{1}\orcidID{0000-0002-6285-1041} \and
Lidia Losavio\inst{1}\orcidID{0000-0002-8834-7388} \and
Giulia Vilone\inst{2}\orcidID{0000-0002-4401-5664} \and
Marc Langheinrich\inst{1}\orcidID{0009-0009-0800-7251}}
\authorrunning{F. Sovrano et al.}
\institute{University of Italian-Speaking Switzerland (USI)\\
\email{\{francesco.sovrano,lidia.anna.maria.losavio,marc.langheinrich\}@usi.ch} \and
Analog Devices International\\
\email{giulia.vilone@analog.com}}

\maketitle
\begin{abstract}
Symbolic regression aims to replace black-box predictors with concise analytical expressions that can be inspected and validated in scientific machine learning.
Kolmogorov--Arnold Networks (KANs) are well suited to this goal because each connection between adjacent units (an \quotes{edge}) is parametrised by a learnable univariate function that can, in principle, be replaced by a symbolic operator.
In practice, however, symbolic extraction is a bottleneck: the standard KAN-to-symbol approach fits operators to each learned edge function in isolation, making the discrete choice sensitive to initialisation and non-convex parameter fitting, and ignoring how local substitutions interact through the full network.
We study \emph{in-context symbolic regression} for operator extraction in KANs, and present two complementary instantiations.
Greedy in-context Symbolic Regression (GSR) performs greedy, in-context selection by choosing edge replacements according to end-to-end loss improvement after brief fine-tuning.
Gated Matching Pursuit (GMP) amortises this in-context selection by training a differentiable gated operator layer that places an operator library behind sparse gates on each edge; after convergence, gates are discretised (optionally followed by a short in-context greedy refinement pass).
We quantify robustness via one-factor-at-a-time (OFAT) hyper-parameter sweeps and assess both predictive error and qualitative consistency of recovered formulas.
Across several experiments, greedy in-context symbolic regression achieves up to 99.8\% reduction in median OFAT test MSE.
\\\textbf{Code \& Data}: \href{https://github.com/Francesco-Sovrano/In-Context-Symbolic-Regression-KAN}{https://github.com/Francesco-Sovrano/In-Context-Symbolic-Regression-KAN}

\end{abstract}

\begin{acronym}
    \acro{XAI}{Explainable AI}
    \acro{MP}{Matching Pursuit}
    \acro{MSE}{Mean Square Error}
    \acro{MLP}{Multilayer Perceptron}
    \acro{KAN}{Kolmogorov--Arnold Network}
    \acro{GMP}{Gated Matching Pursuit}
    \acro{GSR}{Greedy in-context Symbolic Regression}
\end{acronym}

% ======================================================================
\section{Introduction}
% ======================================================================

\ac{XAI} is increasingly expected to support scientific workflows: uncovering functional relationships, proposing compact mechanisms, and producing artefacts that can be inspected and validated \cite{doshivelez2017rigorous,rudin2019stop}. In this setting, \emph{robustness} is essential. If a method produces different analytical expressions for the same dataset under small changes to random seed, hyper-parameters, architecture width, or representation resolution, then its \quotes{explanation} becomes hard to reproduce and difficult to trust \cite{alvarezmelis2018robustness,adebayo2018sanity,ghorbani2019fragile}.

Symbolic regression is attractive as an \emph{intrinsically interpretable} modelling paradigm: it returns an explicit symbolic formula that fits the data, rather than a black-box predictor accompanied by post-hoc explanations \cite{ribeiro2016why,lundberg2017unified,rudin2019stop}. This has driven renewed interest in symbolic regression systems, including modern approaches that combine combinatorial search over expressions with continuous optimisation, simplification, and constant-fitting heuristics \cite{koza1992genetic,schmidt2009distilling,udrescu2020aifeynman,cranmer2023pysr,lacava2021contemporary}.

\acp{KAN} \cite{liu2024kan} offer a promising bridge between neural learning and symbolic formulas, grounded in the Kolmogorov--Arnold superposition view \cite{kolmogorov1957representation,arnold1957functions}. Unlike \acp{MLP}, \acp{KAN} place learnable \emph{univariate} functions on edges and compute node outputs by summation. This design (and extensions for scientific discovery \cite{liu2024kan20}) makes it natural to visualise the learned edge functions, replace them with symbolic operators, and compose them into a final expression.

\begin{figure}[t]
\centering
\includegraphics[width=\linewidth]{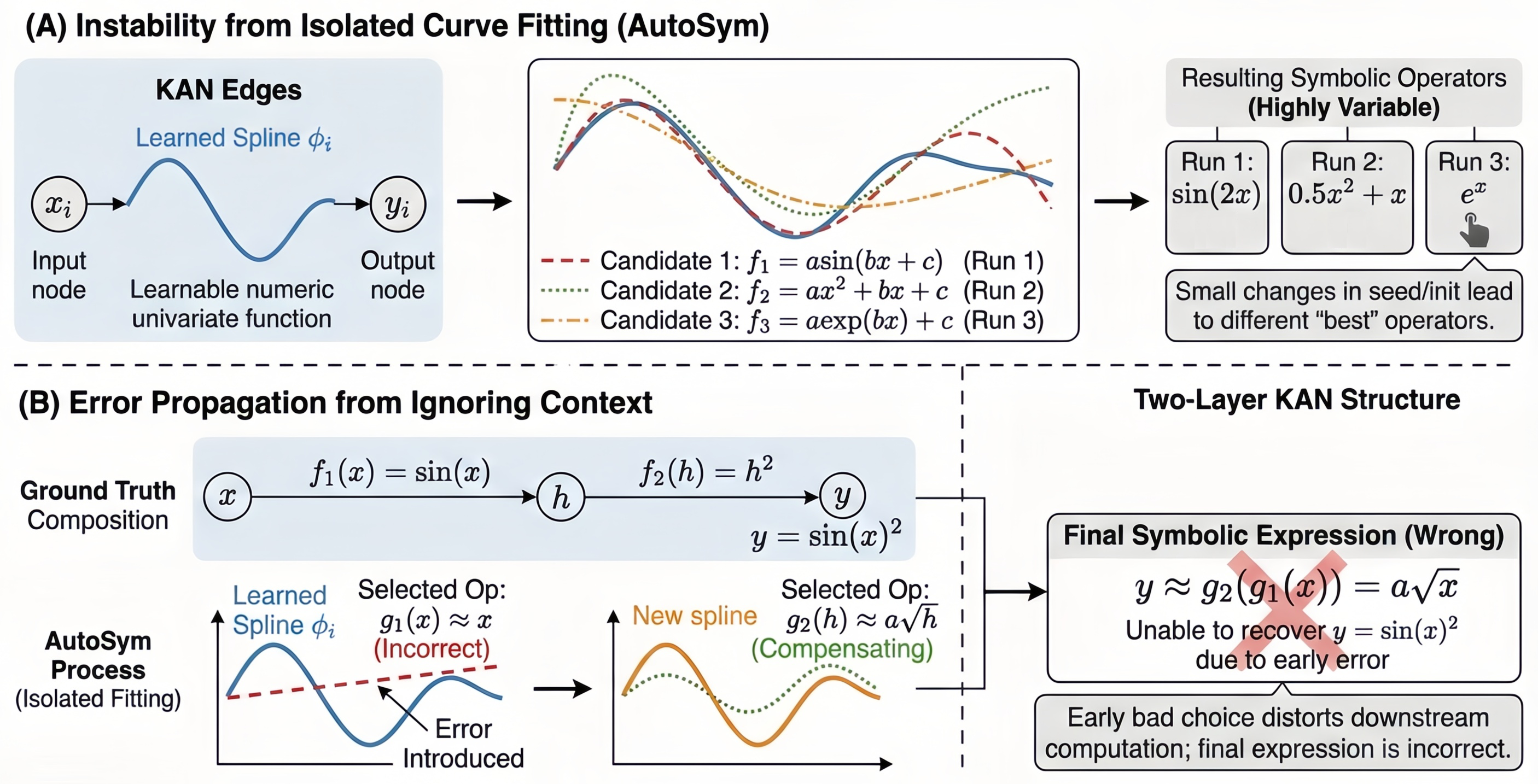}
\caption{Problem overview: isolated per-edge KAN-to-symbol fitting (AutoSym) is unstable and ignores end-to-end context.}
\label{fig:problem_overview}
\end{figure}

In practice, however, symbolic extraction remains a computational and methodological bottleneck: many learned numeric edge functions must be converted into discrete choices from an operator library. A common strategy for \acp{KAN}, which we call \emph{AutoSym}, processes each edge independently. It samples the learned univariate function from its numerical parametrisation (often a spline) and selects the library operator that best fits the sampled curve, optionally with a simplicity penalty \cite{liu2024kan20}. This pipeline is fragile for two reasons:
\begin{enumerate}
  \item \textit{Instability from isolated curve fitting:} When candidate operators include free parameters, e.g., \quotes{$a \cdot \sin(bx+c)+d$}, per-edge fitting is non-convex and sensitive to initialisation and local minima. Moreover, expressive operator families can fit many shapes, so multiple candidates often achieve similar curve-fitting scores, making the choice ambiguous \cite{adebayo2018sanity,ghorbani2019fragile}. As a result, small changes in \ac{KAN} initialisation, spline grid, or training can produce different \quotes{best} operators for the same edge.
  \item \textit{Error propagation from ignoring context:} Even when two operators fit the local spline equally well, their \emph{global} effect inside the full \ac{KAN} can differ. Because AutoSym commits to edge-wise decisions in isolation, it cannot account for interactions between edges. Errors in early symbolic choices may distort downstream computations, forcing later edges to compensate and potentially preventing recovery of the ground-truth composition.
\end{enumerate}
Fig.~\ref{fig:problem_overview} summarises why isolated per-edge extraction is unstable and can propagate errors through the network.

A more stable alternative is to evaluate candidates \emph{in context}. For a given edge, we temporarily replace its numeric function with a candidate operator, briefly fine-tune the \emph{full} network, and score the candidate by the resulting end-to-end loss. We then revert to the pre-trial state and repeat this process for the remaining candidates, finally committing to the operator that yields the largest reduction in global loss after short fine-tuning. Repeating this procedure while prioritising the edge that most improves the objective resembles \ac{MP}, a classic greedy method for sparse approximation that selects atoms from a dictionary to reduce the residual as quickly as possible \cite{mallat1993matching,pati1993omp,tropp2007omp}. Applied to \ac{KAN} symbolic regression, this yields a \ac{GSR} procedure that is substantially more stable than isolated per-edge fitting because it selects operators by their \emph{end-to-end loss improvement}.

The drawback of greedy in-context selection is computational cost, and it still treats symbolic structure as a post-training decision. To address this, we propose \ac{GMP}. The key idea is to replace the numeric edge parametrisation (e.g., a spline basis) with a \textit{gated operator mechanism} that places the entire operator library behind a differentiable gate. During training, the model learns gate weights per edge, effectively performing operator selection \emph{as part of optimisation}. After convergence, we discretise the gates to obtain a single operator per edge and optionally apply a short greedy refinement pass. Our approach is inspired by function-combination mechanisms in \ac{KAN}-like architectures \cite{ta2026fckan} and by mixture-of-experts gating \cite{jacobs1991adaptive,shazeer2017outrageously}. Continuous relaxations for discrete selection (e.g., Gumbel--Softmax) provide practical tools to encourage near-discrete gates during training \cite{jang2017gumbel,maddison2017concrete,louizos2018l0}.

We evaluate \ac{GSR} and \ac{GMP} through the lens of \emph{hyper-parameter robustness}: we sweep random seeds and architectural choices (notably network width, regularisation strength, and pruning schedule) and study how predictive performance and recovered symbolic structure vary across runs. 
Experiments are conducted on \emph{SRBench} \cite{lacava2021contemporary}, a standard benchmark suite for symbolic regression, using tasks from its Feynman collection.
We find that AutoSym is highly sensitive under these sweeps, often producing different operators and final expressions for minor perturbations. In contrast, \emph{in-context} symbolic regression methods (\ac{GSR} and \ac{GMP}) generally exhibit higher hyper-parameter robustness in our experiments, leading to more consistent operator recovery and more stable final formulas; \ac{GSR} is typically the strongest performer, while \ac{GMP} remains competitive.

Overall, this paper makes the following contributions:
\begin{itemize}
  \item We identify failure modes of isolated, per-edge spline-to-symbol fitting for \ac{KAN} symbolic regression and connect them to \ac{XAI} robustness concerns \cite{doshivelez2017rigorous,rudin2019stop,alvarezmelis2018robustness}.
  \item We formalise a Matching-Pursuit-inspired greedy in-context symbolic regression procedure for \acp{KAN} that improves robustness by selecting operators via end-to-end loss after short fine-tuning, thereby reducing error propagation \cite{mallat1993matching,tropp2007omp}.
  \item We introduce a gated operator layer that performs \emph{amortised in-context} operator selection during training, reducing candidate-evaluation cycles; this improves efficiency while retaining much of the robustness benefit of in-context selection \cite{ta2026fckan,shazeer2017outrageously}.
  \item We empirically demonstrate that \emph{in-context} symbolic regression (\ac{GSR}, \ac{GMP}) improves hyper-parameter robustness over isolated per-edge fitting, and often yields lower error with qualitatively more consistent recovered formulas.
  \item We release a working replication package with the code, sweep definitions, and plotting utilities needed to reproduce the experiments, tables, and figures in this paper: \href{https://github.com/Francesco-Sovrano/In-Context-Symbolic-Regression-KAN}{https://github.com/Francesco-Sovrano/In-Context-Symbolic-Regression-KAN}.
\end{itemize}

% ======================================================================
\section{Related Work} \label{sec:related_work}
% ======================================================================

This work connects four lines of research: robustness in \ac{XAI}, symbolic regression, \acp{KAN} and KAN-to-symbol extraction, and greedy/gated selection mechanisms.

\paragraph{Robustness in explainable modelling.}
A central distinction in \ac{XAI} is between \emph{post-hoc} explanations and \emph{intrinsically interpretable} models \cite{doshivelez2017rigorous,rudin2019stop}. Post-hoc methods such as LIME and SHAP provide feature-attribution explanations for arbitrary predictors \cite{ribeiro2016why,lundberg2017unified}, but their outputs can be sensitive to sampling, perturbation schemes, and other implementation choices \cite{alvarezmelis2018robustness,adebayo2018sanity,ghorbani2019fragile}. Prior work therefore emphasises \emph{robustness} as a practical requirement: explanations should not change substantially under small perturbations of training conditions, random seeds, or model specification when predictive behaviour is similar \cite{sovrano2025legal,alvarezmelis2018robustness}.
Symbolic regression produces a model that is itself an explanation. This shifts robustness from a diagnostic property to a core requirement: instability corresponds to recovering different candidate \quotes{laws} from the same data, which undermines reproducibility in scientific use \cite{lacava2021contemporary}.

\paragraph{Symbolic regression.}
Classical symbolic regression traces back to genetic programming and evolutionary search \cite{koza1992genetic}, with early systems demonstrating recovery of compact physical relations from data \cite{schmidt2009distilling}. More recent methods combine discrete search with continuous optimisation and incorporate simplification and constant-fitting heuristics \cite{udrescu2020aifeynman,cranmer2023pysr}. Related directions include sparse model discovery with predefined feature libraries (e.g., SINDy) \cite{brunton2016sindy}. Across these approaches, performance depends on the operator set, noise level, and evaluation protocol, motivating standardised benchmarks such as SRBench \cite{lacava2021contemporary}. In this paper we follow this practice and evaluate on SRBench tasks, with a focus on the hyper-parameter robustness of structure recovery.

\paragraph{\acp{KAN} and symbolic extraction.}
\acp{KAN} were introduced as an alternative to \acp{MLP} in which each edge carries a learnable univariate function (commonly parameterised by splines) and nodes aggregate inputs by summation \cite{liu2024kan}. Follow-up work extended \acp{KAN} toward scientific discovery, including multiplicative variants and tooling for symbolic conversion \cite{liu2024kan20}. Related architectures explore alternative edge parametrisations and basis functions \cite{aghaei2024rkan,ta2026fckan}. FastKAN replaces the B-spline basis with Gaussian radial basis functions and shows that spline-based KAN layers can be approximated by radial basis function networks \cite{li2024kanrbf}; we include FastKAN as a baseline.
A common use case is to train a numeric \ac{KAN} and then convert it to a closed-form expression by fitting each learned edge function to an operator library and composing the resulting operators \cite{liu2024kan20}. As discussed in our introduction, this per-edge extraction is local and can be sensitive to non-convex parameter fitting and to interactions between edges, motivating methods that evaluate operators in context and/or integrate selection into training.

\paragraph{Greedy pursuit for in-context selection.}
\ac{MP} is a greedy procedure for sparse approximation that iteratively selects dictionary atoms to reduce the residual \cite{mallat1993matching}. Variants such as Orthogonal Matching Pursuit provide improved recovery in certain regimes \cite{pati1993omp,tropp2007omp}. We adapt the same ``select-and-refine'' principle to \ac{KAN} symbolic extraction: candidates are scored by end-to-end loss after brief fine-tuning, and selections are committed iteratively. This in-context selection directly targets the global objective, rather than relying on isolated curve fitting.

\paragraph{Gating mechanisms and continuous relaxations.}
Mixture-of-experts models use gating networks to select or weight expert components \cite{jacobs1991adaptive}, with sparsely gated variants enabling efficient scaling by activating only a subset of experts per input \cite{shazeer2017outrageously}. Discrete selection can be approximated with continuous relaxations such as Gumbel--Softmax \cite{jang2017gumbel,maddison2017concrete}, and sparsity can be encouraged via regularisation (e.g., $\ell_0$-style penalties) \cite{louizos2018l0}. We use gating at the level of \ac{KAN} edges: each edge maintains a gated mixture over a symbolic operator library during training, which is later discretised to obtain a single operator per edge.

% ======================================================================
\section{Background} \label{sec:background}
% ======================================================================

This section introduces the notation and components used in the proposed methods: the \ac{KAN} layer, the MultKAN extension, the standard per-edge symbolic extraction baseline, and \ac{MP}, which serves as the greedy template underlying \ac{GSR} and the refinement stage of \ac{GMP}.

\paragraph{\ac{KAN} layers.}
The Kolmogorov--Arnold representation theorem states that continuous multivariate functions on compact domains can be expressed using compositions of univariate functions and addition \cite{kolmogorov1957representation,arnold1957functions}. \acp{KAN} instantiate this idea by parameterising each edge with a learnable univariate function \cite{liu2024kan}.
Consider a layer mapping $x\in\mathbb{R}^{d_{\mathrm{in}}}$ to $y\in\mathbb{R}^{d_{\mathrm{out}}}$. A \ac{KAN} layer computes
\begin{equation}
  y_j \;=\; \sum_{i=1}^{d_{\mathrm{in}}} \phi_{j,i}(x_i),
  \label{eq:kan-layer}
\end{equation}
where $\phi_{j,i}:\mathbb{R}\rightarrow\mathbb{R}$ is the learnable 1D function associated with edge $(i\!\to\! j)$. In the original formulation, each $\phi_{j,i}$ is represented by a spline basis \cite{liu2024kan,deboor1978splines}. Stacking layers yields compositions of these edge functions across depth.

\paragraph{MultKAN.}
Many scientific expressions involve explicit products. MultKAN extends \acp{KAN} by adding multiplication modules (or multiplication nodes) so that multiplicative interactions are represented directly \cite{liu2024kan20}. Our methods apply to both \ac{KAN} and MultKAN; unless otherwise stated, we use the additive \ac{KAN} notation in Eq.\ref{eq:kan-layer}.

\paragraph{Per-edge symbolic extraction baseline.}
Let $\phi_{j,i}$ denote a trained edge function represented numerically (e.g., by spline coefficients). The standard KAN-to-symbol baseline approximates $\phi_{j,i}$ by selecting an operator family from a library:
\[
  \phi_{j,i}(x) \approx g_k(x;\theta),
\]
where $g_k$ is the $k$-th candidate operator and $\theta$ are its continuous parameters (e.g., scale/shift/frequency).

\paragraph{Examples of operator families.}
A typical library contains a set of univariate primitives with affine reparametrisations. For example:
\begin{align*}
  g_{\sin}(x;\theta) &= a\,\sin(bx+c)+d, \\
  g_{\exp}(x;\theta) &= a\,\exp(bx+c)+d, \\
  g_{\log}(x;\theta) &= a\,\log(bx+c)+d \quad (\text{with } bx+c>0),
\end{align*}
where $\theta$ collects the corresponding coefficients (i.e., $\theta=(a,b,c,d)$). Here, $g_k$ denotes the symbolic \emph{form} (e.g., ``sine''), while $\theta$ captures the fitted continuous parameters for that form.

\paragraph{Baseline fitting procedure.}
Given samples $\{(x_t,\phi_{j,i}(x_t))\}_{t=1}^T$ from the numeric edge function, the baseline fits $\theta$ for each candidate family by minimising a local regression loss, e.g.
\[
  \min_{\theta}\;\frac{1}{T}\sum_{t=1}^{T}\left(\phi_{j,i}(x_t)-g_k(x_t;\theta)\right)^2,
\]
and then selects the candidate $k$ with the best local score, i.e., lowest \ac{MSE} or highest $R^2$, optionally with a simplicity penalty \cite{liu2024kan20}. In the experiments reported here we set that simplicity weight to zero. The reason is technical: on the bounded domains induced by the training data, several operator families can fit the same sampled edge almost equally well after affine reparametrisation, so a hand-assigned complexity score can dominate near-ties for reasons unrelated to end-to-end fidelity. Because such scores depend on the chosen library and are not invariant to algebraically equivalent representations, they can systematically steer AutoSym toward a cheaper-but-wrong family. We therefore disable this heuristic to isolate the effect of local-versus-in-context evaluation.

\paragraph{Baseline Limitations.}
Per-edge fitting is often non-convex in $\theta$, and several operator families can fit the same sampled curve comparably well under affine reparametrisations, e.g., $a \cdot \sin(bx+c)+d$. As a result, the selected operator can depend on initialisation and optimisation details. Moreover, a good \emph{local} fit does not guarantee that replacing $\phi_{j,i}$ by $g_k(\cdot;\theta)$ preserves end-to-end performance once the edge is inserted back into the full network.

\paragraph{Matching Pursuit.}
\ac{MP} is a greedy method for building a sparse approximation from a fixed library (dictionary) of candidate components \cite{mallat1993matching}. Starting from an initial approximation, it repeatedly (i) selects the single candidate that yields the largest improvement according to a criterion, and (ii) updates the approximation before proceeding to the next selection. Orthogonal variants such as \textit{OMP} modify the update step to re-estimate coefficients after each selection \cite{pati1993omp,tropp2007omp}.
In this paper, \ac{MP} is used as an algorithmic template rather than a signal-processing tool. The ``dictionary'' is the symbolic operator library, and candidates are instantiated by assigning an operator family to a specific edge (with its parameters fitted). The improvement criterion is not residual norm, but \emph{end-to-end loss improvement} of the whole \ac{KAN}. This is why \ac{GSR} evaluates a candidate by inserting it into the network, fitting its continuous parameters (and, if needed, allowing a short end-to-end re-fit of affected parameters), and measuring the resulting loss. The same select--update structure also motivates the optional greedy refinement pass applied after gate discretisation in \ac{GMP}.

% ======================================================================
\section{Proposed Methods} \label{sec:proposed_methods}
% ======================================================================

\begin{figure}[t]
\centering
\includegraphics[width=\linewidth]{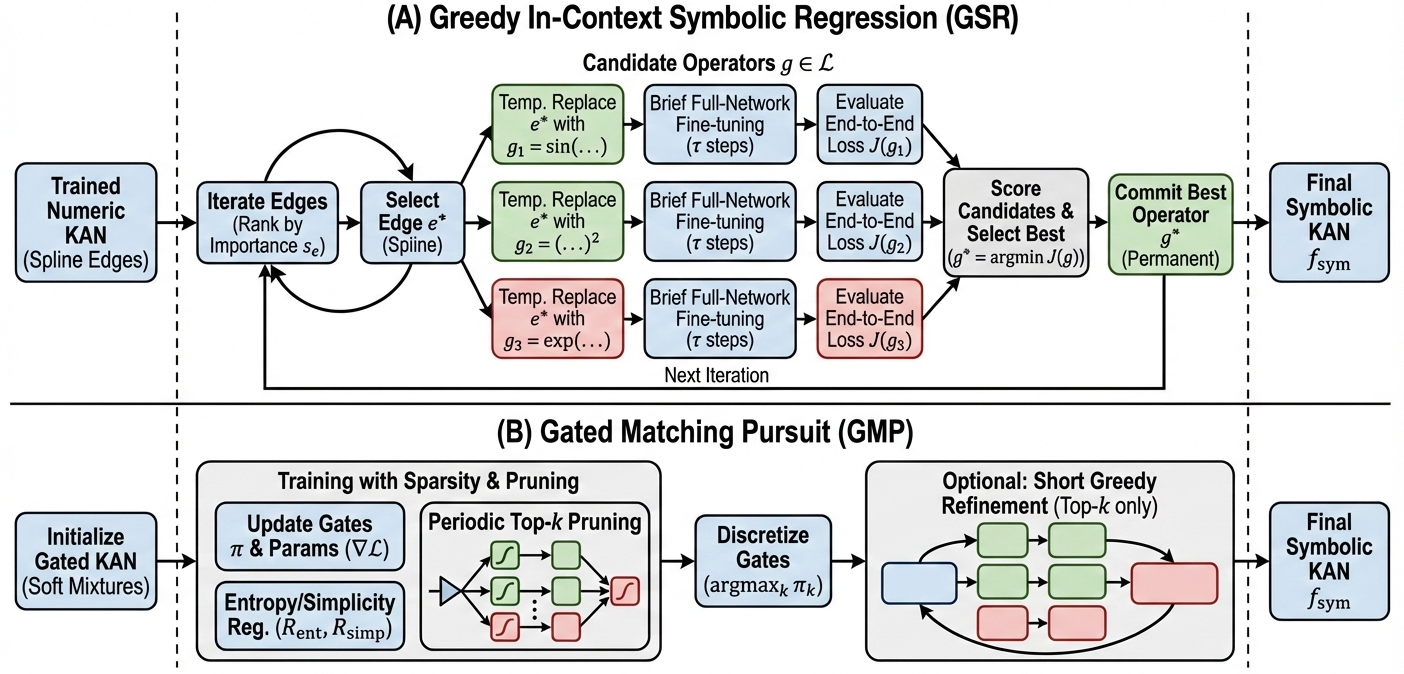}
\caption{Method overview: \ac{GSR} selects operators by end-to-end loss improvement; \ac{GMP} amortises in-context selection via sparse operator gates during training, then discretises (optionally refined by a short greedy pass) to reduce candidate-trial cost.}
\label{fig:methods_overview}
\end{figure}

We aim to turn a trained numeric \ac{KAN} whose edges are spline functions into an interpretable \emph{symbolic} \ac{KAN} whose edges come from a small operator library (e.g., $\sin$, polynomials, $\exp$).
We propose two complementary conversion strategies.
\ac{GSR} is a post-hoc procedure that converts one edge at a time by \emph{trying candidate operators, briefly fine-tuning the whole network, and committing the operator that yields the best end-to-end loss}.
\ac{GMP} accelerates this search by learning \emph{soft operator gates} during training, pruning each edge to a small top-$k$ candidate set, discretising, and optionally running a short greedy refinement restricted to those candidates.
Fig.~\ref{fig:methods_overview} summarises the workflow.

\subsection{Problem formulation}
We start from a trained numeric \ac{KAN} (or MultKAN) model $f_{\text{num}}$ whose edge functions $\phi^{\text{num}}_{j,i}:\mathbb{R}\!\to\!\mathbb{R}$ are represented numerically (in practice, splines; cf. Eq.~\ref{eq:kan-layer}). We seek a functionally similar but more interpretable model $f_{\text{sym}}$ obtained by replacing a subset of these spline edges with closed-form operators.

Let $\mathcal{L}=\{g_k\}_{k=1}^{K}$ be a library of univariate operator \emph{forms} (e.g., $\sin$, $\exp$, $\log$, polynomials; see Section~\ref{sec:background}).
When an edge $e=(i\!\to\! j)$ is converted, we represent it as
\begin{equation}
  \phi^{\text{sym}}_{j,i}(x)
  \;=\;
  \alpha_{e}\, g_{k_e}(\beta_{e} x + \gamma_{e})
  + \delta_{e},
  \label{eq:symbolic-edge}
\end{equation}
where $k_e\in\{1,\dots,K\}$ selects the operator form and $(\alpha_e,\beta_e,\gamma_e,\delta_e)$ are learnable affine parameters that absorb scale/shift.

We call an edge \emph{numeric} if it is still spline-based, and \emph{symbolic} once it takes the form in Eq.~\ref{eq:symbolic-edge}.
Our objective is to obtain $f_{\text{sym}}$ that preserves predictive performance on held-out data (e.g., $\mathcal{J}(f_{\text{sym}};\mathcal{D}_{\mathrm{val}})\le \mathcal{J}(f_{\text{num}};\mathcal{D}_{\mathrm{val}})+\varepsilon$), while favouring stable operator choices across seeds/hyper-parameters and an efficient conversion budget. When a domain-specific complexity prior is well defined it can be added on top of this objective, but we do not enforce such a prior in the experiments of this paper.

\paragraph{Running example.}
If the ground-truth is $y=\sin(x_1)+x_2^2$, a trained numeric \ac{KAN} typically learns spline edges that approximate $\sin(\cdot)$ and $(\cdot)^2$ on the data range. Our methods replace those spline edges with the corresponding operator forms from $\mathcal{L}$, while briefly re-optimising the full network to account for interactions between edges.

\subsection{\acl{GSR}} \label{sec:proposed_methods:gsr}

\paragraph{Edge ranking (existing \ac{KAN} heuristic).}
At each iteration, we prioritise which \emph{numeric} edge to convert using an importance score $s_e$ (as commonly used for pruning in \acp{KAN}). \emph{Greedily}, we select the single remaining numeric edge with the highest score,
$e^\star = \arg\max_{e\ \text{numeric}} s_e$,
so early conversions target the most influential edges. In our implementation, $s_e$ is recomputed from the current network state before each selection (optionally amortised by updating only after committing a batch of edges).

In each iteration, \ac{GSR} selects the most important remaining \emph{numeric} edge $e^\star$ (according to the importance score $s_e$), then evaluates candidate operators in-context. Concretely, for each $g\in\mathcal{L}$ (or a pruned subset) we temporarily replace $e^\star$ with $g$ (including affine parameters), fine-tune the \emph{full network} for $\tau$ steps, measure the resulting end-to-end loss on a small validation split, and restore the original parameters. We finally commit the operator that yields the lowest loss and proceed to the next edge.

This procedure, formally represented as Algorithm~\ref{alg:gsr}, mitigates error propagation because each symbolic choice is evaluated in the context of all previously committed symbolic choices and the remaining numeric edges. It also mitigates initialisation sensitivity: rather than selecting the operator that best fits a spline in isolation (where local minima dominate), we select the operator that yields the best global objective after brief adaptation. 

\begin{algorithm}[t]
\caption{\acl{GSR} for \acp{KAN}}
\label{alg:gsr}
\scriptsize
\begin{algorithmic}[1]
\REQUIRE Trained numeric \ac{KAN} $f$, dataset $\mathcal{D}$, operator library $\mathcal{L}$, fine-tuning steps $\tau$, max symbolic edges $M$
\ENSURE Symbolic \ac{KAN} model $f_\text{sym}$
\STATE Compute initial edge importance scores $\{s_e\}$ for all edges $e$
\STATE Initialize set of converted edges $\mathcal{S}\leftarrow\emptyset$
\FOR{$m=1$ to $M$}
  \STATE Recompute/update edge importance scores $\{s_e\}$ (optional)
  \STATE Select next edge $e^\star \leftarrow \arg\max_{e\notin\mathcal{S}} s_e$
  \STATE Initialize best loss $J^\star \leftarrow +\infty$ and best operator $g^\star \leftarrow \text{None}$
  \FOR{each candidate operator $g \in \mathcal{L}$ (or a pruned subset)}
    \STATE Snapshot model parameters (including all edges)
    \STATE Replace edge $e^\star$ with operator $g$ (initialise/retain affine params)
    \STATE Fine-tune full model for $\tau$ steps on $\mathcal{D}$
    \STATE Evaluate end-to-end loss $J \leftarrow \mathcal{J}(f;\mathcal{D}_{\mathrm{val}})$
    \STATE Restore snapshot
    \IF{$J < J^\star$}
      \STATE $J^\star \leftarrow J$, $g^\star \leftarrow g$
    \ENDIF
  \ENDFOR
  \STATE Commit: replace edge $e^\star$ with $g^\star$ permanently
  \STATE Optional: brief fine-tuning after committing
  \STATE Update $\mathcal{S} \leftarrow \mathcal{S}\cup\{e^\star\}$
\ENDFOR
\STATE \textbf{return} $f$ as $f_\text{sym}$
\end{algorithmic}
\end{algorithm}

\paragraph{Complexity of naive \ac{GSR}.}
Let $K=|\mathcal{L}|$ be the library size and $\tau$ be the fine-tuning budget per trial. Naively, each symbolic selection costs $K$ trial runs, each with $\tau$ optimisation steps, yielding $\mathcal{O}(K\tau)$ steps per converted edge. For $M$ edges, this is $\mathcal{O}(MK\tau)$ optimisation steps, often unacceptable in practice.

\subsection{Gated operator layers for amortised in-context selection} \label{sec:proposed_methods:gmp}

To reduce the number of costly in-context trial runs required by \ac{GSR}, we integrate operator selection directly into training by introducing \emph{gated operator layers}. This yields \ac{GMP}, an \emph{amortised} variant of in-context selection: rather than explicitly trying every operator for every edge, each edge maintains a differentiable mixture over the operator library, and training learns both (i) which operator(s) to prefer and (ii) the corresponding continuous parameters. After training, we discretise the gates to obtain a single symbolic operator per edge and optionally apply a short greedy refinement pass restricted to the retained candidates.

\paragraph{Gated edge parametrisation.}
For each edge $(i\!\rightarrow\! j)$, instead of a spline $\phi_{j,i}$ we use a soft selection over the operator library $\mathcal{L}=\{g_k\}_{k=1}^{K}$:
\begin{equation}
  \phi_{j,i}(x)
  \;=\;
  \sum_{k=1}^{K}
    \pi^{(j,i)}_{k}\;
    \Big(
      \alpha^{(j,i)}_{k}\, g_k(\beta^{(j,i)}_{k} x + \gamma^{(j,i)}_{k})
      + \delta^{(j,i)}_{k}
    \Big),
  \label{eq:gated-edge}
\end{equation}
where $\pi^{(j,i)} \in \Delta^{K-1}$ is a probability vector over operators and
$(\alpha,\beta,\gamma,\delta)$ are per-operator affine parameters that absorb scale/shift, matching the symbolic edge form in Eq.~\ref{eq:symbolic-edge}.

We parameterise the gate via logits $\ell^{(j,i)} \in \mathbb{R}^K$ and a softmax:
\begin{equation}
  \pi^{(j,i)}_{k}
  \;=\;
  \frac{\exp(\ell^{(j,i)}_k)}{\sum_{r=1}^K \exp(\ell^{(j,i)}_r)}.
  \label{eq:softmax-gate}
\end{equation}
Intuitively, each edge carries a mixture of candidate symbolic operators, and optimisation increases the weight of operators that reduce end-to-end loss \emph{in context} (because the mixture is trained as part of the full network).

\paragraph{Stabilising the gate with variance compression.}
Different operators can produce outputs with very different scales and heavy-tailed responses, which can destabilise optimisation and make the logits $\ell^{(j,i)}$ overly sensitive to outliers. To mitigate this, we compress each operator output $z$ with a \emph{scaled asinh} transform before mixing:
\begin{equation}
  \tilde z
  \;=\;
  s\,\operatorname{asinh}\!\Big(\frac{z}{s}\Big),
  \label{eq:compress-asinh-scaled}
\end{equation}
where the scale parameter $\log s$ is learned jointly with
$(\ell,\alpha,\beta,\gamma,\delta)$ (optionally per edge/operator). This transform is approximately linear for $|z|\ll s$ and grows only logarithmically for $|z|\gg s$, damping extreme values while preserving small variations.

\paragraph{Encouraging sparsity and enabling discretisation.}
To obtain a final symbolic model, each edge should select (approximately) a single operator. We encourage near-discrete gates with two complementary heuristics:
(i) entropy regularisation,
$R_{\mathrm{ent}}=\sum_{(j,i)} H(\pi^{(j,i)})$,
to favour peaky distributions; and
(ii) periodic top-$k$ pruning, which keeps only the $k$ highest-probability operators per edge and masks out the rest.

\paragraph{Training, pruning, and discretisation.}
\ac{GMP} trains the gated \ac{KAN} end-to-end by minimising
\[
J(\cdot; D) + \lambda_{\mathrm{ent}} R_{\mathrm{ent}}
+ \lambda_{\ell_1} R_{\ell_1},
\]
where $R_{\ell_1}$ is an $\ell_1$ penalty on the gate parameters. During training, we periodically prune each edge to its top-$k$ operators according to $\pi^{(j,i)}$. A hand-crafted simplicity term could be added here in principle, but we keep it disabled in all reported experiments for the same identifiability reasons discussed for AutoSym above. After training, we discretise each edge by taking $\arg\max_k \pi^{(j,i)}_{k}$, replacing the mixture in Eq.~\ref{eq:gated-edge} with that single operator and retaining its learned affine parameters.

\paragraph{Optional greedy refinement (restricted \ac{GSR}).}
Although gating already performs in-context selection during training, discretisation is a hard decision and may occasionally pick between near-ties. To validate and potentially correct these discrete choices, we optionally run a short greedy refinement pass using \ac{GSR} (Algorithm~\ref{alg:gsr}) but restricting each edge’s candidate set to its retained top-$k$ operators (as determined by pruning). This preserves the computational benefits of \ac{GMP} while adding a targeted in-context check at the end.

\begin{algorithm}[t]
\caption{\acl{GMP}}
\label{alg:gmp}
\scriptsize
\begin{algorithmic}[1]
\REQUIRE Dataset $\mathcal{D}$, operator library $\mathcal{L}$, gated \ac{KAN} architecture, training steps $T$, pruning schedule, top-$k$ value $k$, refinement steps $\tau$
\ENSURE Symbolic \ac{KAN} $f_\text{sym}$
\STATE Initialize gated \ac{KAN}: each edge is a mixture over $\mathcal{L}$ (Eq.~\ref{eq:gated-edge})
\FOR{$t=1$ to $T$}
  \STATE Update network parameters by minimizing $\mathcal{J}(\cdot;\mathcal{D}) + \lambda_{\mathrm{ent}} R_{\mathrm{ent}} + \lambda_{\ell_1} R_{\ell_1}$
  \IF{pruning step}
    \STATE For each edge, keep top-$k$ operators by gate probability; mask out the rest
  \ENDIF
\ENDFOR
\STATE Discretize: for each edge, choose $g^\star = \arg\max_{g\in\mathcal{L}} \pi(g)$; replace mixture with $g^\star$
\STATE Optional refinement: run \ac{GSR} (Alg.~\ref{alg:gsr}) but restrict each edge's candidate set to its retained top-$k$
\STATE \textbf{return} symbolic model $f_\text{sym}$
\end{algorithmic}
\end{algorithm}

\paragraph{Why \ac{GMP} should improve robustness over isolated per-edge fitting.}
\ac{GMP} avoids isolated spline-to-operator curve fitting and instead learns operator preferences \emph{jointly} with the full network objective, reducing sensitivity to local minima and per-edge initialisation.
Sparsity objectives and top-$k$ pruning further reduce ambiguity among similarly fitting operators and help prevent oscillation.
Finally, the optional restricted \ac{GSR} refinement evaluates discrete operator choices explicitly in context, correcting occasional discretisation errors at low additional cost.

% ======================================================================
\section{Experiments} \label{sec:experiments}
% ======================================================================

We evaluate our methods with a focus on \emph{robustness}: whether small, routine changes in training and conversion settings lead to large changes in predictive performance and in the recovered symbolic model. %We additionally report end-to-end runtime, since in-context operator evaluation is the dominant cost driver.

\subsection{Tasks and data protocol}

We use regression datasets from the SRBench \emph{Feynman} benchmark \cite{lacava2021contemporary}, a standard benchmark suite for symbolic regression. We specifically chose the Feynman \emph{with-units} suite because it provides controlled scientific symbolic-regression tasks with known closed-form targets, making it suitable for studying symbolic recovery in \acp{KAN}. Within that suite, we evaluate 10 targets (I.10.7, I.12.1, I.13.4, II.34.29a, I.9.18, I.12.4, I.34.1, II.6.15a, II.6.15b, and II.21.32), selected without further per-problem tuning so as to keep the OFAT study computationally tractable while reducing cherry-picking. Each dataset contains multiple input variables and a single scalar target.

% We use datasets from \emph{SRBench} \cite{lacava2021contemporary}, a standard benchmark suite for symbolic regression; specifically, we evaluate 10 randomly selected targets from its \emph{Feynman with-units} collection: I.10.7, I.12.1, I.13.4, II.34.29a, I.9.18, I.12.4, I.34.1, II.6.15a, II.6.15b, and II.21.32. Each dataset contains multiple input variables and a single scalar target.

For each dataset, we construct a training and test split by sampling up to 2000 training points and 1000 test points from the available rows (when a dataset contains fewer rows, we use the maximum available under these caps). Unless otherwise stated, the split is obtained by a seeded random permutation of rows. Predictive performance is measured by test \ac{MSE}.

\subsection{Compared pipelines}
All methods use the same univariate operator library $\mathcal{L}$ with $K=25$ operator forms:
constants and identity; polynomial powers $x^2$--$x^5$; inverse powers $1/x$--$1/x^3$; $\sqrt{\cdot}$ and $1/\sqrt{\cdot}$;
$\log$ and $\exp$; $\sin,\cos,\tan,\tanh$; $|\cdot|$ and $\mathrm{sgn}$; $\arctan$; $\arcsin$; $\arccos$; $\mathrm{arctanh}$; and a Gaussian primitive $\exp(-x^2)$.
Each symbolic edge additionally includes a learnable affine reparametrisation as in Eq.~\ref{eq:symbolic-edge}.

Using a relatively large library makes operator selection substantially more challenging; in particular, it increases the combinatorial ambiguity faced by post-hoc selection and makes \ac{GMP} harder to optimise, since learning sparse and confident gates becomes more complex as $K$ grows.

We compare five pipelines:
\begin{enumerate}
  \item \textbf{AutoSym (baseline).}
  Train a numeric MultKAN \cite{liu2024kan20} whose edges are spline functions, then replace each remaining active edge independently by fitting candidate operators to its learned one-dimensional curve and selecting the best local fit. In our experiments we disable any explicit complexity bias in this local selection to isolate the effect of in-context evaluation (simplicity weight $=0$).

  \item \textbf{FastKAN + AutoSym.}
  Same post-hoc per-edge extraction as AutoSym, but the numeric edge parametrisation uses radial basis functions instead of splines \cite{li2024kanrbf}.

  \item \textbf{\ac{GSR} (cf. Section \ref{sec:proposed_methods:gsr}).}
  Train the same numeric model as AutoSym, then perform greedy in-context symbolic regression: iteratively choose one still-numeric edge using the model's edge-importance scores, try candidate operators on that edge, briefly refit the full model, and commit the operator that yields the lowest loss after this short refit.

  \item \textbf{FastKAN + \ac{GSR} (cf. Section \ref{sec:proposed_methods:gsr}).}
  Same greedy in-context conversion as \ac{GSR}, starting from a radial-basis numeric parametrisation \cite{li2024kanrbf}.

  \item \textbf{\ac{GMP} (cf. Section \ref{sec:proposed_methods:gmp}).}
  Train a model whose edges are differentiable gated mixtures over the operator library, with gate sparsity encouraged by entropy regularisation and periodic top-$k$ pruning. During pruning rounds, each edge is restricted to a small shortlist of operators according to its gate weights. After training, we discretise each edge by selecting the operator with the largest gate probability and retaining its learned affine parameters. We then optionally apply a short restricted greedy refinement pass, using GSR only over each edge's retained top-$k$ candidates. This yields an efficiency-oriented in-context baseline that reduces candidate-trial cost while preserving much of the robustness benefit of in-context selection.
\end{enumerate}

\subsection{Training schedule, sensitivity sweep, and reported metrics}

\paragraph{Model family and training schedule.}
All methods share the same base architecture: a single hidden MultKAN layer followed by a scalar output. The hidden layer contains $m$ additive units and \emph{two} multiplication units, i.e.\ width $[m,2]$, where $m$ is varied in the sweep.
For spline-based models, inputs are mapped onto a fixed grid resolution of {20} knots/centres (depending on the numeric parametrisation) and we use {cubic B-splines} (degree $3$). The grid range is set per dataset to the minimum and maximum observed in the training inputs (global min/max across all input dimensions).

Each run follows the same multi-stage schedule: an initial fit without regularisation, followed by several prune-and-refit cycles with regularisation enabled, then a final non-regularised fit. Symbolic extraction (AutoSym or greedy conversion) is applied after this final fit, and we perform a short final polishing fit afterwards.
Training uses the \textit{Adam} optimizer with learning rate $10^{-2}$. Each fit stage uses a fixed budget of {200} optimisation steps. Greedy candidate evaluations use a budget of {100} steps per candidate. During prune-and-refit cycles, pruning uses a node threshold of {0.1}; edge-threshold pruning is disabled (edge threshold $=0.0$). The regularised refit uses the same optimiser and step budget as the non-regularised stages, differing only by the regularisation coefficient $\lambda$ (below).

For \ac{GMP}, gate sparsity is encouraged with an entropy penalty weight of {$10^{-3}$} and an $\ell_1$ gate penalty weight of {$10^{-2}$}. Gate pruning uses an initial cap of {10} operators per edge and decreases to the final shortlist size (top-$k$) of {5} across pruning cycles.
In all reported GMP experiments, we enabled the optional restricted-GSR refinement after gate discretisation; this refinement used the same short candidate-evaluation budget as GSR and was restricted to each edge’s retained top-$k$ shortlist.

% For \ac{GSR}, the candidate-evaluation fine-tuning horizon $\tau$ is kept short to reflect a practical in-context trial budget.
% For \ac{GMP}, pruning is applied after an initial warm-up phase, gates are discretised at convergence, and (when enabled) greedy refinement is restricted to the retained top-$k$ candidates per edge.

\paragraph{One-factor-at-a-time sensitivity sweep.}
We operationalise robustness as low sensitivity of predictive performance and recovered formulas to routine experimental perturbations. To measure this sensitivity, we use a one-factor-at-a-time sweep around a reference configuration. For each dataset we vary exactly one factor at a time while holding the others fixed:
\begin{itemize}
  \item \textbf{Hidden width:} number of additive units $m\in\{5,10,20,50,100\}$ (multiplication units fixed to $2$).
  \item \textbf{Regularisation strength $\lambda$:} $\{10^{-4},10^{-3},10^{-2},10^{-1}\}$ during prune-and-refit cycles.
  \item \textbf{Number of pruning cycles:} $\{1,3,5\}$.
  \item \textbf{Random seed:} $\{1,2,3\}$, controlling initialisation and the random train-test split.
\end{itemize}
The \textit{reference configuration} is $(m,\lambda,\text{\#cycles},\text{seed})=(5,10^{-2},3,1)$. This is the default anchor setting used in three places: it is the unperturbed point around which each OFAT sweep is constructed, it is the configuration repeated across the factor-specific sweeps, and it is the fixed setting used for the seed-only repeats reported in Table~\ref{tab:feynman_seed_sensitivity}. Counting repeated appearances of that anchor yields 15 runs per dataset (12 unique configurations), and we run all five pipelines for each run.

\paragraph{Metrics.}
For each method and dataset we report:
\begin{itemize}
  \item \textbf{Predictive accuracy:} test \ac{MSE}.
  \item \textbf{Sensitivity-based robustness proxy:} the distribution of test \ac{MSE} over the sweep, summarised by the median (and dispersion via quartiles). Lower, tighter distributions indicate greater robustness to the perturbed factor.
\end{itemize}
% We also record the recovered symbolic expression for each run to support qualitative inspection of structural consistency across the sweep.

\paragraph{Statistical comparison.}
For each dataset, we select the method with the lowest median test \ac{MSE} as the reference.
We then compare the reference to each other method using a one-sided Mann--Whitney U test with alternative hypothesis
$\text{MSE(ref)} < \text{MSE(other)}$.
We correct for multiple comparisons (reference vs.\ each competitor) using Holm correction.
To complement significance testing, we report effect size using Cliff's $\Delta$ with 95\% bootstrap confidence intervals.

% ======================================================================
\section{Results}
% ======================================================================

We report two complementary views of sensitivity, which together support our robustness claims. First, Table~\ref{tab:feynman_seed_sensitivity} gives a \emph{seed-sensitivity snapshot} at the fixed reference configuration, summarised as mean$\pm$std over repeated seeds. Second, Figure~\ref{fig:ofat_violins} visualises the \emph{OFAT hyper-parameter sensitivity distributions} obtained by varying width, $\lambda$, and the number of pruning cycles around that same reference configuration; Table~\ref{tab:mwu_ofat_sensitivity_comparison} summarises their median performance relative to the AutoSym baseline, and Table~\ref{tab:mwu_ofat_sensitivity} reports the corresponding distribution-level statistical comparisons. Figure~\ref{fig:ofat_violins} is therefore \emph{not} an average over the seed repeats in Table~\ref{tab:feynman_seed_sensitivity}, and Table~\ref{tab:feynman_seed_sensitivity} is \emph{not} an average of the points shown in Figure~\ref{fig:ofat_violins}. The rankings can differ because the two summaries answer different questions.

% ----------------------------------------------------------------------
\subsection{Seed sensitivity at the reference configuration}
% ----------------------------------------------------------------------

Table~\ref{tab:feynman_seed_sensitivity} reports test MSE as mean$\pm$std over available seeds for the reference configuration only; this is a seed-sensitivity snapshot rather than an OFAT summary. Entries marked with $^\dagger$ use fewer than three successful seeds, and some GMP runs are unavailable under the default settings. 
Across the 10 datasets, the lowest mean test MSE at this fixed setting is achieved by FastKAN+GSR on I.10.7, I.12.1, I.12.4, and I.34.1; by GSR on I.13.4, I.9.18, and II.6.15b; by FastKAN+AutoSym on II.34.29a and II.21.32; and by GMP on II.6.15a, although that last result is based on fewer successful seeds and should be interpreted cautiously. 
These fixed-setting means are not directly comparable to the OFAT medians in Figure~\ref{fig:ofat_violins}: the latter aggregate all valid configurations in the sweep, whereas Table~\ref{tab:feynman_seed_sensitivity} averages only repeated seeds of one selected setting.

The missing or limited GMP entries are consistent with a schedule-mismatch failure mode under the shared training protocol: gated operator layers can separate more slowly than the greedy post-hoc variants, so pruning may remove viable operator paths before the gates have stabilised. We therefore treat these cases as procedural failures under the current schedule rather than as definitive evidence that GMP cannot fit the task; a longer pre-pruning phase or a milder pruning schedule may recover some of them. Overall, the in-context greedy pipelines (GSR and FastKAN+GSR) are the strongest and most consistently competitive at the reference setting, while several post-hoc pipelines exhibit substantial stochastic variability on selected targets. Importantly, the standard deviations reveal substantial stochastic variability for some pipelines and datasets (e.g., AutoSym on I.13.4), motivating the broader OFAT robustness analysis below.

\begin{table}[t]
\centering
\caption{Seed sensitivity at the reference configuration (width $[5,2]$, $\lambda=10^{-2}$, and three pruning cycles). We report test MSE as mean$\pm$std over available seeds for this fixed setting only. This table quantifies sensitivity to stochasticity at one chosen configuration; lower means and smaller standard deviations indicate greater robustness, but it is not an average of the OFAT sweep in Figure~\ref{fig:ofat_violins}. Entries marked with $^\dagger$ use fewer than three successful seeds; for single-seed cases, only the observed value is reported. Entries marked N/A indicate that no successful run was obtained under the shared training/pruning schedule. Lower is better. The lowest mean is in \textbf{bold}; the second-lowest mean is \underline{underlined}.}
\label{tab:feynman_seed_sensitivity}
\resizebox{0.95\columnwidth}{!}{%
\begin{tabular}{p{.12\linewidth}|cc|cc|c}
\toprule
Feynman Dataset & AutoSym & FastKAN+AutoSym & GSR & FastKAN+GSR & GMP \\
\midrule
I.13.4
& 1.31e3 $\pm$ 2.21e3
& 1.29e1 $\pm$ 1.77e1
& \textbf{7.35e-1 $\pm$ 3.94e-1}
& \underline{1.42e0 $\pm$ 6.61e-1}
& N/A \\

I.10.7
& 6.14e-2 $\pm$ 1.38e-2
& 8.49e-1 $\pm$ 8.64e-1
& \underline{2.60e-3 $\pm$ 2.50e-3}
& \textbf{1.90e-3 $\pm$ 3.00e-3}
& 1.14e-2 $\pm$ 4.20e-3 \\

I.12.1
& 1.07e2 $\pm$ 1.43e2
& 4.86e-1 $\pm$ 4.79e-1
& \underline{7.18e-2 $\pm$ 2.25e-2}
& \textbf{1.30e-2 $\pm$ 1.31e-2}
& 1.25e0 $\pm$ 9.99e-1 \\

II.34.29a
& 7.18e-4 $\pm$ 1.23e-4
& \textbf{8.60e-5 $\pm$ 1.07e-4}
& 6.43e-4 $\pm$ 3.50e-4
& \underline{5.99e-4 $\pm$ 3.77e-4}
& 1.18e-2 $\pm$ 1.38e-2 \\

I.9.18
& 1.40e-1 $\pm$ 2.15e-1
& 6.05e-3 $\pm$ 9.40e-3
& \textbf{2.89e-4 $\pm$ 1.53e-4}
& \underline{1.57e-3 $\pm$ 2.01e-3}
& N/A \\

I.12.4
& 4.64e-4 $\pm$ 3.70e-4
& \underline{1.12e-4 $\pm$ 1.86e-4}
& 1.24e-4 $\pm$ 2.50e-5
& \textbf{8.20e-5 $\pm$ 6.60e-5}
& 1.73e-4 $\pm$ 3.60e-5 \\

I.34.1
& 1.27e-1 $\pm$ 1.03e-1
& 3.39e-1 $\pm$ 4.75e-1
& \underline{2.02e-2 $\pm$ 3.16e-2}
& \textbf{8.77e-3 $\pm$ 1.20e-2}
& 3.53e-2 $\pm$ 2.39e-2 \\

II.6.15a
& 1.28e-1 $\pm$ 1.12e-1
& 4.51e0 $\pm$ 7.81e0
& 3.13e-3 $\pm$ 2.33e-3
& \underline{2.96e-3 $\pm$ 1.19e-3}
& \textbf{2.22e-3$^{\dagger}$} \\

II.6.15b
& 7.67e-4 $\pm$ 2.12e-4
& \underline{3.86e-4 $\pm$ 4.29e-4}
& \textbf{3.25e-4 $\pm$ 2.28e-4}
& 5.00e-4 $\pm$ 3.53e-4
& 8.33e-4 $\pm$ 9.80e-5 \\

II.21.32
& 1.15e-3 $\pm$ 1.50e-3$^{\dagger}$
& \textbf{1.00e-5 $\pm$ 4.00e-6$^{\dagger}$}
& \underline{7.80e-5 $\pm$ 5.20e-5}
& 8.80e-5 $\pm$ 4.00e-5
& 1.02e-3$^{\dagger}$ \\
\bottomrule
\end{tabular}%
}
\end{table}

\subsection{Hyper-parameter sensitivity under OFAT sweeps}

\begin{figure}%[H]
\centering

  \includegraphics[width=\linewidth]{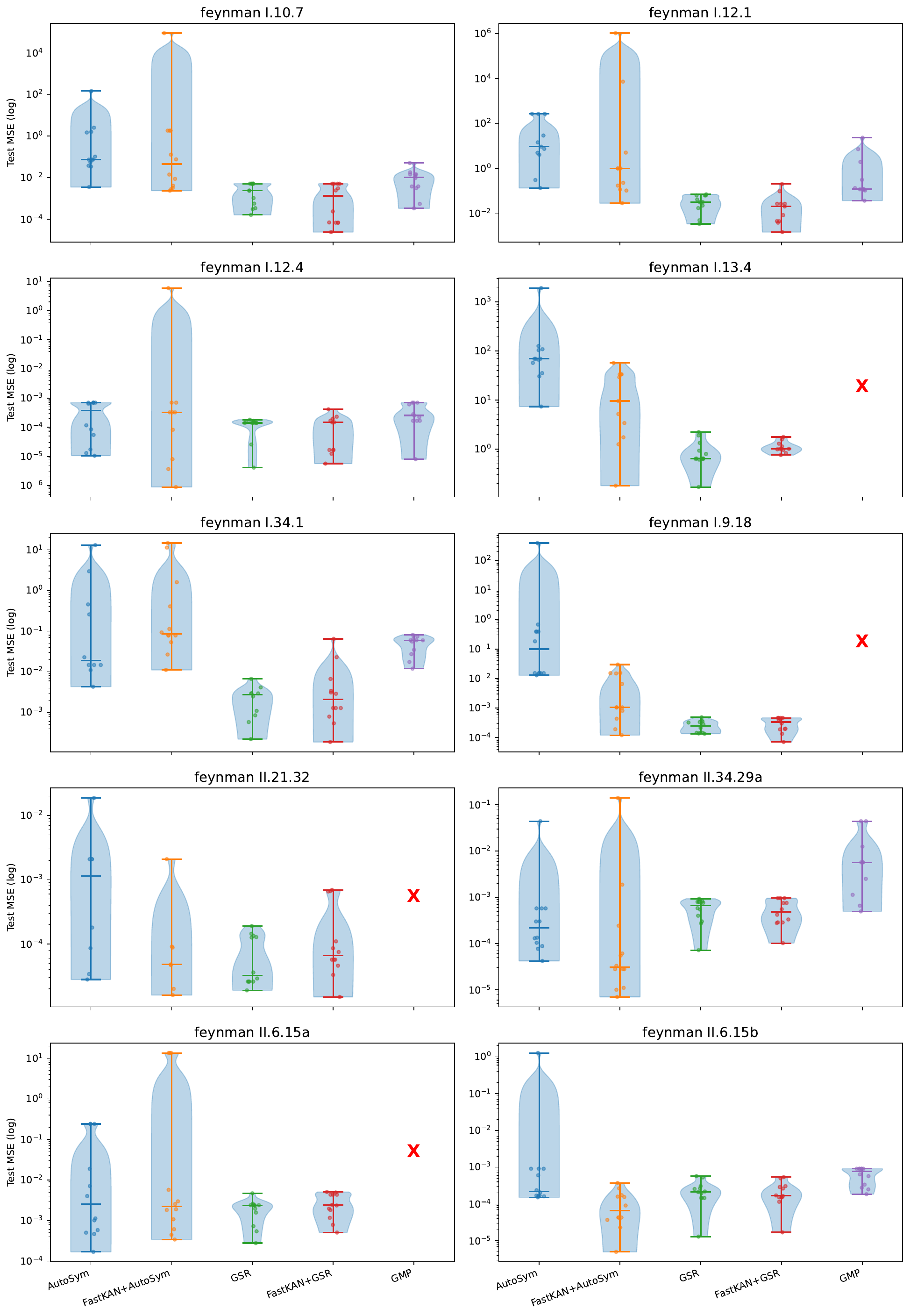}
  \caption{OFAT hyper-parameter sensitivity distributions. Violins summarise test MSE across all valid one-factor-at-a-time runs obtained by varying hidden width, $\lambda$, and the number of pruning cycles around the reference configuration; dots denote individual observations. This figure aggregates hyper-parameter perturbations only and does not average over the seed-only repeats from Table~\ref{tab:feynman_seed_sensitivity}. Red $\times$ markers indicate that a method produced no valid OFAT runs for that dataset and is therefore absent from the violin aggregation. Lower, tighter distributions indicate lower sensitivity and hence greater robustness.}
\label{fig:ofat_violins}
\end{figure}

We analyse sensitivity to hyper-parameters by aggregating each method's test MSE values across the OFAT sweep dimensions of hidden width, $\lambda$, and the number of pruning cycles (we exclude the explicit \quotes{seed} factor here, since seed sensitivity is already reported in Table~\ref{tab:feynman_seed_sensitivity}).
Figure~\ref{fig:ofat_violins} shows the resulting \emph{hyper-parameter sensitivity distributions} as violin plots on a log-scale y-axis (necessary due to the wide dynamic range of MSE).
Lower medians and tighter distributions indicate that a method is less sensitive to routine hyper-parameter choices and therefore more robust under this operationalisation.
Some methods, especially GMP on selected datasets, produced no valid OFAT observations under the shared pruning schedule; these cases are shown as explicit missing-value markers rather than being silently dropped.

% \paragraph{Statistical comparison.}
We report distribution-level comparisons of OFAT robustness using one-sided Mann--Whitney $U$ tests with Holm correction (details in Section~\ref{sec:experiments}).
In Table~\ref{tab:mwu_ofat_sensitivity}, $p_{\mathrm{Holm}}$ indicates significance after correction, Cliff's $\delta$ quantifies the effect size (more negative favours the reference), and the 95\% bootstrap CI for \quotes{$\mathrm{median(other)}-\mathrm{median(ref)}$} reports the median gap (positive favours the reference).
Table~\ref{tab:mwu_ofat_sensitivity} summarises these results and highlights which comparisons remain significant after Holm correction.
Across datasets, the statistical conclusions are consistent with the distributions in Fig.~\ref{fig:ofat_violins}.
%
% For interpretability, note that the largest median improvement over the AutoSym baseline occurs on Feynman I.12.1:
For interpretability, note that among the 10 datasets the largest median improvement over the AutoSym baseline occurs on Feynman I.12.1:
FastKAN+GSR reduces the median OFAT test MSE from 9.49 to 0.0212, i.e.,
a $100\cdot(1-0.0212/9.49)\approx 99.8\%$ reduction (Table~\ref{tab:mwu_ofat_sensitivity_comparison}).

\begin{table}
\centering
\caption{Median OFAT test MSE of the best pipeline vs.\ AutoSym baseline, and percent reduction $100\cdot(1-\mathrm{med(best)}/\mathrm{med(AutoSym)})$.}
\label{tab:mwu_ofat_sensitivity_comparison}
\resizebox{0.4\linewidth}{!}
{\begin{tabular}{p{.13\linewidth}lrr} 
\toprule
\begin{tabular}[c]{@{}l@{}}Feynman~\\Dataset\end{tabular} & \begin{tabular}[c]{@{}l@{}}med\\(best)\end{tabular} & \multicolumn{1}{c}{\begin{tabular}[c]{@{}c@{}}med\\(AutoSym)\end{tabular}} & \multicolumn{1}{c}{\begin{tabular}[c]{@{}c@{}}Reduction\\(\%)\end{tabular}} \\ 
\midrule
I.10.7 & 1.34e-3 & 7.36e-2 & 98.2 \\
I.12.1 & 2.12e-2 & 9.49e0 & 99.8 \\
I.13.4 & 6.43e-1 & 6.94e1 & 99.1 \\
II.34.29a & 3.05e-5 & 2.17e-4 & 85.9 \\
II.21.32 & 3.25e-5 & 1.13e-3 & 97.1 \\
II.6.15a & 2.23e-3 & 2.58e-3 & 13.6 \\
II.6.15b & 6.70e-5 & 2.22e-4 & 69.8 \\
I.12.4 & 1.41e-4 & 3.81e-4 & 63.0 \\
I.34.1 & 2.09e-3 & 1.88e-2 & 88.9 \\
I.9.18 & 2.48e-4 & 9.92e-2 & 99.8 \\
\bottomrule
\end{tabular}}
\end{table}

% On Feynman I.10.7 and I.12.1, \texttt{FastKAN+GSR} attains the lowest median OFAT test MSE and significantly outperforms \texttt{AutoSym} and \texttt{FastKAN+AutoSym}; its advantage over \texttt{GSR} is not significant after correction, suggesting that the two greedy in-context pipelines exhibit comparable robustness on these targets.
% On Feynman I.13.4, \texttt{GSR} yields the lowest median OFAT test MSE and significantly outperforms \texttt{AutoSym}, \texttt{FastKAN+AutoSym}, and \texttt{FastKAN+GSR}.
% On Feynman II.34.29a, \texttt{FastKAN+AutoSym} achieves the lowest median OFAT test MSE and significantly outperforms the remaining pipelines (including the greedy in-context variants), indicating that for this target the radial-basis numeric parametrisation combined with post-hoc extraction is particularly robust under the tested sweep.

Across the 10 datasets, an in-context variant attains the lowest median OFAT test MSE on seven targets: FastKAN+GSR on I.10.7, I.12.1, and I.34.1, and GSR on I.13.4, II.21.32, I.12.4, and I.9.18. FastKAN+AutoSym attains the lowest OFAT median on the remaining three targets: II.34.29a, II.6.15a, and II.6.15b. These OFAT median rankings do not always match the fixed-setting mean rankings in Table~\ref{tab:feynman_seed_sensitivity}, because the OFAT analysis aggregates all valid settings in the sweep whereas Table~\ref{tab:feynman_seed_sensitivity} summarises only repeated seeds of one selected configuration. The contrast is clearest on II.21.32: FastKAN+AutoSym has the lowest mean at the reference configuration in Table~\ref{tab:feynman_seed_sensitivity}, but GSR has the lowest median across the full OFAT sweep in Figure~\ref{fig:ofat_violins} and Table~\ref{tab:mwu_ofat_sensitivity_comparison}; Table~\ref{tab:mwu_ofat_sensitivity} further shows that the GSR vs FastKAN+AutoSym comparison is not Holm-significant on this dataset. In the clearest cases favouring in-context selection (namely I.10.7, I.12.1, I.13.4, I.9.18, and I.34.1), the best greedy variant significantly outperforms AutoSym after Holm correction, often with large effect sizes. At the same time, the wins of FastKAN+AutoSym on II.34.29a, II.6.15a, and II.6.15b indicate that the underlying numeric parametrisation can in some cases make local post-hoc extraction sufficiently robust. Other datasets are less clear-cut. For example, on I.12.4, GSR achieves the best OFAT median, but after correction only the comparison with GMP remains significant. On II.6.15a, none of the pairwise differences remains significant after correction despite a small median advantage for FastKAN+AutoSym.

\begin{table}%[H]
\centering
\caption{One-sided Mann--Whitney $U$ tests comparing the best pipeline (lowest median OFAT test MSE) to the others. $p$-values are Holm-corrected per dataset; effect size is Cliff's $\delta$ (negative favours the best pipeline). Bootstrap 95\% CIs are reported for the median difference (other$-$best). Only methods with valid OFAT distributions are included; omitted rows correspond to unavailable distributions (notably selected GMP cases). {Significance markers:} * for $p_{\mathrm{Holm}}<0.05$, ** for $p_{\mathrm{Holm}}<0.01$, and *** for $p_{\mathrm{Holm}}<0.001$.}
\label{tab:mwu_ofat_sensitivity}
\resizebox{0.95\columnwidth}{!}{%
\begin{tabular}{lp{.1\linewidth}p{.1\linewidth}p{.12\linewidth}rr}
\toprule
Comparison ($\text{best} \rightarrow \text{other}$) & \begin{tabular}[c]{@{}r@{}}med\\(best)\end{tabular} & \begin{tabular}[c]{@{}r@{}}med\\(other)\end{tabular} & $p_{\mathrm{Holm}}$ & Cliff's $\delta$ & CI \\
\midrule
\multicolumn{6}{l}{\textbf{Feynman I.10.7} (best: FastKAN+GSR)} \\
FastKAN+GSR $\rightarrow$ AutoSym (baseline) & 1.34e-3 & 7.36e-2 & 1.89e-4\sym{\textbf{***}} & -0.944 & [4.51e-2, 1.55e0] \\
FastKAN+GSR $\rightarrow$ FastKAN+AutoSym & 1.34e-3 & 4.52e-2 & 3.59e-3\sym{\textbf{**}} & -0.736 & [1.35e-3, 1.84e0] \\
FastKAN+GSR $\rightarrow$ GMP & 1.34e-3 & 1.06e-2 & 6.01e-3\sym{\textbf{**}} & -0.667 & [-2.83e-4, 1.43e-2] \\
FastKAN+GSR $\rightarrow$ GSR & 1.34e-3 & 2.38e-3 & 6.27e-2 & -0.375 & [-3.39e-3, 5.05e-3] \\
\addlinespace

\multicolumn{6}{l}{\textbf{Feynman I.12.1} (best: FastKAN+GSR)} \\
FastKAN+GSR $\rightarrow$ AutoSym (baseline) & 2.12e-2 & 9.49e0 & 2.06e-4\sym{\textbf{***}} & -0.983 & [4.14e0, 2.72e2] \\
FastKAN+GSR $\rightarrow$ FastKAN+AutoSym & 2.12e-2 & 1.03e0 & 4.42e-4\sym{\textbf{***}} & -0.917 & [9.89e-2, 5.20e0] \\
FastKAN+GSR $\rightarrow$ GMP & 2.12e-2 & 1.23e-1 & 6.21e-4\sym{\textbf{***}} & -0.868 & [9.25e-2, 1.98e0] \\
FastKAN+GSR $\rightarrow$ GSR & 2.12e-2 & 3.31e-2 & 1.18e-1 & -0.306 & [-4.62e-3, 6.06e-2] \\
\addlinespace

\multicolumn{6}{l}{\textbf{Feynman I.13.4} (best: GSR)} \\
GSR $\rightarrow$ AutoSym (baseline) & 6.43e-1 & 6.94e1 & 5.32e-5\sym{\textbf{***}} & -1.000 & [4.57e1, 1.06e2] \\
GSR $\rightarrow$ FastKAN+AutoSym & 6.43e-1 & 9.59e0 & 2.27e-3\sym{\textbf{**}} & -0.758 & [1.10e0, 3.28e1] \\
GSR $\rightarrow$ FastKAN+GSR & 6.43e-1 & 1.02e0 & 2.60e-2\sym{\textbf{*}} & -0.485 & [-1.22e-1, 6.52e-1] \\
\addlinespace

\multicolumn{6}{l}{\textbf{Feynman II.34.29a} (best: FastKAN+AutoSym)} \\
FastKAN+AutoSym $\rightarrow$ AutoSym (baseline) & 3.05e-5 & 2.17e-4 & 1.07e-2\sym{\textbf{*}} & -0.569 & [2.94e-5, 5.48e-4] \\
FastKAN+AutoSym $\rightarrow$ FastKAN+GSR & 3.05e-5 & 4.87e-4 & 1.07e-2\sym{\textbf{*}} & -0.653 & [2.28e-4, 8.29e-4] \\
FastKAN+AutoSym $\rightarrow$ GMP & 3.05e-5 & 5.72e-3 & 4.23e-3\sym{\textbf{**}} & -0.783 & [1.07e-3, 2.50e-2] \\
FastKAN+AutoSym $\rightarrow$ GSR & 3.05e-5 & 6.72e-4 & 1.07e-2\sym{\textbf{*}} & -0.653 & [3.03e-4, 7.67e-4] \\
\addlinespace

\multicolumn{6}{l}{\textbf{Feynman II.21.32} (best: GSR)} \\
GSR $\rightarrow$ AutoSym (baseline) & 3.25e-5 & 1.13e-3 & 4.06e-2\sym{\textbf{*}} & -0.604 & [-4.10e-5, 2.06e-3] \\
GSR $\rightarrow$ FastKAN+AutoSym & 3.25e-5 & 4.80e-5 & 5.51e-1 & 0.0238 & [-8.70e-5, 6.50e-5] \\
GSR $\rightarrow$ FastKAN+GSR & 3.25e-5 & 6.60e-5 & 2.84e-1 & -0.264 & [-7.30e-5, 3.48e-4] \\
\addlinespace

\multicolumn{6}{l}{\textbf{Feynman II.6.15a} (best: FastKAN+AutoSym)} \\
FastKAN+AutoSym $\rightarrow$ AutoSym (baseline) & 2.23e-3 & 2.58e-3 & 1.00e0 & 0.0833 & [-6.77e0, 1.28e-1] \\
FastKAN+AutoSym $\rightarrow$ FastKAN+GSR & 2.23e-3 & 2.39e-3 & 1.00e0 & 0.0139 & [-6.77e0, 2.88e-3] \\
FastKAN+AutoSym $\rightarrow$ GSR & 2.23e-3 & 2.37e-3 & 1.00e0 & 0.167 & [-6.77e0, 1.32e-3] \\
\addlinespace

\multicolumn{6}{l}{\textbf{Feynman II.6.15b} (best: FastKAN+AutoSym)} \\
FastKAN+AutoSym $\rightarrow$ AutoSym (baseline) & 6.70e-5 & 2.22e-4 & 5.24e-3\sym{\textbf{**}} & -0.708 & [1.30e-5, 8.67e-4] \\
FastKAN+AutoSym $\rightarrow$ FastKAN+GSR & 6.70e-5 & 1.68e-4 & 4.60e-2\sym{\textbf{*}} & -0.431 & [-5.00e-6, 2.46e-4] \\
FastKAN+AutoSym $\rightarrow$ GMP & 6.70e-5 & 7.76e-4 & 2.73e-4\sym{\textbf{***}} & -0.917 & [2.13e-4, 8.67e-4] \\
FastKAN+AutoSym $\rightarrow$ GSR & 6.70e-5 & 2.16e-4 & 4.60e-2\sym{\textbf{*}} & -0.486 & [1.45e-5, 2.25e-4] \\
\addlinespace

\multicolumn{6}{l}{\textbf{Feynman I.12.4} (best: GSR)} \\
GSR $\rightarrow$ AutoSym (baseline) & 1.41e-4 & 3.81e-4 & 6.23e-1 & -0.125 & [-1.05e-4, 5.59e-4] \\
GSR $\rightarrow$ FastKAN+AutoSym & 1.41e-4 & 3.27e-4 & 5.07e-1 & -0.250 & [-1.35e-4, 5.59e-4] \\
GSR $\rightarrow$ FastKAN+GSR & 1.41e-4 & 1.47e-4 & 6.23e-1 & 0.000 & [-1.36e-4, 5.62e-5] \\
GSR $\rightarrow$ GMP & 1.41e-4 & 2.54e-4 & 2.96e-3\sym{\textbf{**}} & -0.788 & [1.59e-5, 5.59e-4] \\
\addlinespace

\multicolumn{6}{l}{\textbf{Feynman I.34.1} (best: FastKAN+GSR)} \\
FastKAN+GSR $\rightarrow$ AutoSym (baseline) & 2.09e-3 & 1.88e-2 & 2.12e-3\sym{\textbf{**}} & -0.783 & [9.68e-3, 1.61e0] \\
FastKAN+GSR $\rightarrow$ FastKAN+AutoSym & 2.09e-3 & 8.57e-2 & 1.89e-4\sym{\textbf{***}} & -0.944 & [6.28e-2, 9.99e-1] \\
FastKAN+GSR $\rightarrow$ GMP & 2.09e-3 & 5.96e-2 & 9.31e-4\sym{\textbf{***}} & -0.848 & [2.49e-2, 6.28e-2] \\
FastKAN+GSR $\rightarrow$ GSR & 2.09e-3 & 2.72e-3 & 6.42e-1 & 0.0833 & [-3.26e-3, 1.91e-3] \\
\addlinespace

\multicolumn{6}{l}{\textbf{Feynman I.9.18} (best: GSR)} \\
GSR $\rightarrow$ AutoSym (baseline) & 2.48e-4 & 9.92e-2 & 5.24e-5\sym{\textbf{***}} & -1.000 & [1.49e-2, 3.89e-1] \\
GSR $\rightarrow$ FastKAN+AutoSym & 2.48e-4 & 1.05e-3 & 2.90e-3\sym{\textbf{**}} & -0.722 & [3.90e-4, 1.48e-2] \\
GSR $\rightarrow$ FastKAN+GSR & 2.48e-4 & 3.37e-4 & 1.85e-1 & -0.222 & [-1.30e-4, 2.89e-4] \\
\bottomrule
\end{tabular}%
}
\end{table}

% ======================================================================
\section{Discussion} \label{sec:discussion}

This section interprets the robustness patterns observed in the seed-sensitivity snapshot (Table~\ref{tab:feynman_seed_sensitivity}) and in the OFAT sensitivity distributions (Fig.~\ref{fig:ofat_violins}), as well as the distribution-level comparisons (Table~\ref{tab:mwu_ofat_sensitivity}). We focus on how these results relate to robustness under routine experimental choices.

\paragraph{Interpreting OFAT violins as sensitivity signals.}
The OFAT violin plots in Fig.~\ref{fig:ofat_violins} provide a compact summary of hyper-parameter sensitivity.
Since each violin represents the distribution of test MSE obtained by varying one factor at a time (width, $\lambda$, and the number of pruning cycles) around the reference configuration, the vertical extent of a violin (i.e., the spread of MSE values on the log scale) can be read as an empirical proxy for robustness to routine hyper-parameter choices: taller violins indicate larger performance variability, while flatter violins indicate greater robustness.
This interpretation is consistent with our robustness objective, because it reflects how strongly the pipeline's predictive behaviour changes under small, standard tuning decisions.

\paragraph{Implications of in-context selection and numeric parametrisation.}
Taken together, the multi-seed snapshot at the reference setting (Table~\ref{tab:feynman_seed_sensitivity}) and the OFAT sensitivity distributions (Fig.~\ref{fig:ofat_violins}), supported by distribution-level tests (Table~\ref{tab:mwu_ofat_sensitivity}), motivate three empirical observations:

(i) In the datasets where an in-context variant achieves the lowest median OFAT MSE (GSR or FastKAN+GSR), the corresponding OFAT violins are concentrated at lower error levels and typically exhibit reduced dispersion relative to post-hoc edge-wise extraction.
(ii) Switching the numeric edge parametrisation from splines to radial basis functions can improve robustness for certain targets (e.g., II.34.29a), but it does not uniformly eliminate the variability induced by local per-edge fitting across all datasets.
(iii) When in-context variants outperform post-hoc baselines with Holm-significant differences, the accompanying effect sizes are typically large (Cliff's $\delta$ close to $-1$), indicating that the performance advantage is not limited to a small median shift but reflects a broad separation between the OFAT distributions; the bootstrap intervals for \quotes{$\mathrm{median(other)}-\mathrm{median(best)}$} are predominantly positive in these cases, supporting that the median gaps are practically meaningful across the sweep.

\paragraph{Why in-context evaluation improves robustness.}
A central failure mode of isolated per-edge extraction is that a locally good one-dimensional fit can be globally wrong once composed through additions and multiplication units.
By selecting operators based on the \emph{end-to-end} loss after brief refitting, \ac{GSR} (and the discretisation stage of \ac{GMP}) directly tests whether a symbolic substitution remains compatible with the rest of the network.
This \quotes{in-context} check tends to reject operators that match an edge curve in isolation but introduce brittle interactions downstream, which is consistent with the tighter OFAT sensitivity distributions observed for the greedy pipelines on most targets, with an in-context variant achieving the best median OFAT MSE on seven of the ten datasets.

\paragraph{Why the best pipeline can be target-dependent.}
FastKAN+AutoSym is best by median on three datasets: II.34.29a, II.6.15a, and II.6.15b. This suggests that for some targets the radial-basis parametrisation yields edge functions that are easier to discretize reliably with local post-hoc matching. By contrast, GSR or FastKAN+GSR is best on the remaining seven datasets. End-to-end in-context evaluation is therefore generally more robust, but not universally dominant. More broadly, these results suggest that robustness is shaped jointly by the \emph{numeric} inductive bias (how edges are parametrized and trained) and the \emph{symbolic} selection rule used for discretization.

\paragraph{Efficiency--robustness trade-off.}
While \ac{GSR} can be the most accurate and robust, it incurs a higher conversion cost because it evaluates many candidate operators in context.
\ac{GMP} reduces this cost by learning sparse gates and restricting in-context evaluation to a per-edge top-$k$ shortlist, which directly reduces the number of candidate fine-tuning loops.
The mixed results across targets indicate that gate learning can be an effective accelerator, but that part of GMP's mixed robustness may be procedural rather than intrinsic: the gated operator layer can converge more slowly than the greedy baselines, so applying the same prune-and-refit cadence can prune viable operator paths before gate separation stabilises. This offers a plausible explanation for the missing or invalid GMP runs in Table~\ref{tab:feynman_seed_sensitivity} and Fig.~\ref{fig:ofat_violins}, and suggests that longer pre-pruning training or gentler pruning schedules could improve the efficiency--robustness trade-off.

% ======================================================================
\section{Limitations} \label{sec:limitations}
% ======================================================================

% \fixme{What's below should be double-checked and shortened}

Our study is a first step toward \emph{robustness-aware} symbolic extraction for \acp{KAN}. Below we outline key limitations and, for each, the concrete steps we took to mitigate its impact in the present work, together with what remains open.

\paragraph{Benchmark scope and ecological validity.}
We evaluate on ten targets from SRBench's Feynman suite under a fixed training protocol. This controlled setting does not span the full range of regimes relevant to scientific discovery (e.g., heavy label noise, covariate shift, sparse observations, or high-dimensional inputs), so the results should be interpreted as evidence about robustness \emph{within} this envelope rather than as a general guarantee.
\emph{Mitigation.} We fixed the evaluation pool to the SRBench Feynman \emph{with-units} suite, which offers controlled scientific targets with known formulas, and then used a limited subset without per-problem retuning to reduce cherry-picking under our compute budget.

\paragraph{Limited robustness factors and interaction effects.}
Our robustness analysis performs one-factor-at-a-time sweeps over width, regularisation strength, and pruning rounds, plus a limited set of random seeds. This isolates individual sensitivities but does not fully characterise higher-order interactions (e.g., particular width--pruning combinations that jointly trigger collapse). A larger factorial design or targeted interaction probes (e.g., conditional sweeps around failure regions) would provide stronger guarantees at higher computational cost.
\emph{Mitigation.} We (i) sweep the most consequential knobs for \acp{KAN} extraction (capacity, sparsification, and pruning), (ii) replicate runs across multiple seeds to reduce the risk of anecdotal conclusions, and (iii) report variability across configurations rather than only best-case points.

% \paragraph{Schedule mismatch for gated training.}
% The current protocol uses the same stage length and pruning cadence for all pipelines. This keeps comparisons controlled, but it may disadvantage GMP because gated operator mixtures can require more optimization steps before the gate weights separate cleanly enough for pruning to be safe. When pruning is applied too early, viable operator paths may be removed, leading to failed runs or artificially weak post-discretization performance.
% \emph{Mitigation.} We keep the schedule fixed across methods so that the comparison is not confounded by pipeline-specific retuning, and we make the resulting missing GMP cases explicit rather than hiding them in the aggregate summaries.

\paragraph{Operator-library design and identifiability.}
Operator recovery is constrained by the library and by representational equivalences: distinct operators (or affine reparametrisations) can be indistinguishable on a bounded domain, and mathematically equivalent expressions can appear in different syntactic forms after discretisation and simplification. Consequently, \quotes{correct recovery} is not always uniquely defined without an explicit equivalence protocol.
\emph{Mitigation.} We keep the operator set fixed across pipelines, evaluate candidates on the same input domain, and treat expression recovery primarily as a robustness/interpretability signal rather than as a strict exact-match objective. We also apply consistent post-hoc simplification so that superficial syntactic differences are reduced when reporting recovered formulas.
% \emph{Remaining gap.} A principled equivalence-checking protocol (e.g., algebraic normalization plus domain-aware verification) and domain-specific constraints (units/dimensions, invariances) would be necessary for definitive structural claims.

% \paragraph{Computational cost and scaling.}
% In-context selection is more expensive than isolated per-edge curve fitting because each candidate operator is judged by end-to-end loss after short fine-tuning. Scaling to larger libraries, deeper networks, or more demanding datasets could therefore become costly.
% \emph{Mitigation.} We introduce \ac{GMP} to restrict evaluation to a small per-edge top-$k$ shortlist, and we use short, local fine-tuning rather than full retraining for every candidate. In addition, structured pruning reduces the number of active edges that require operator decisions, which directly lowers the search space.
% \emph{Remaining gap.} Further improvements likely require better amortization (e.g., shared warm-starts, caching, or learned proposal distributions over operators) and more scalable training of sparse gates for large libraries.

\paragraph{What is optimised vs.\ what is explained.}
Our primary quantitative metric is test \ac{MSE}. Predictive accuracy is necessary but not sufficient for high-quality explanations: users often care about structural correctness, robustness of the extracted form under perturbations, and faithfulness in a causal/functional sense beyond numeric fit.
\emph{Mitigation.} We complement \ac{MSE} with qualitative inspection of recovered expressions and, crucially, with a \emph{robustness-oriented} evaluation: we stress the pipelines under controlled hyper-parameter and pruning perturbations to assess whether the extracted forms persist or degrade.
This directly targets one dimension of explanation reliability (robustness to plausible training variations), even when a formal structural metric is unavailable.
The study is therefore best read as comparing pipeline robustness under a shared practical protocol, not as isolating the effect of in-context selection under a strictly matched compute budget.

% ======================================================================
\section{Conclusion and Future Work} \label{sec:conclusions}

We studied symbolic extraction for \acp{KAN} from a \ac{XAI} perspective, emphasising \emph{robustness} under routine experimental choices and operationalising it through sensitivity to controlled perturbations.
We introduced two in-context pipelines that select symbolic operators based on their end-to-end effect on the network after brief fine-tuning: \acl{GSR}, which evaluates candidates explicitly during conversion, and \acl{GMP}, which amortises much of this selection by learning sparse operator gates and then discretising from a small top-$k$ shortlist.

Across ten Feynman targets, the results indicate that in-context selection can substantially increase hyper-parameter robustness relative to isolated per-edge curve matching.
On seven of the ten targets, an in-context variant attains the best OFAT median test \ac{MSE}, often with tighter sensitivity distributions, suggesting improved robustness to width, regularisation, and pruning schedules.
We also find that changing the numeric parametrisation (splines vs.\ radial basis functions) can shift which pipeline is most robust, with FastKAN+AutoSym performing best on II.34.29a, II.6.15a, and II.6.15b under the OFAT median criterion. This highlights that robustness is a property of the \emph{full pipeline} rather than of symbolic extraction alone.

% \paragraph{Future work.}
A few directions for future work could improve robustness and practical usefulness. One is broader evaluation, both on additional SRBench suites and on real scientific datasets, especially in noisy, low-sample, and out-of-distribution regimes where robustness matters most. Another is to go beyond test error by incorporating measures of structural agreement, such as operator-set overlap, edit distance over expression trees, or equivalence-class scoring, and by examining interaction effects through more systematic sweep designs. It would also be worth studying tighter relaxations for operator selection, including $L_0$/Concrete gates, hierarchical operator libraries, and adaptive operator generation, to reduce ambiguity without substantially increasing compute.

\begin{credits}
\subsubsection{\ackname}
This work was funded by the Swiss Innovation Agency (Innosuisse) under grant agreement 119.321 INT-ICT.

% \subsubsection{\discintname} 
% The authors have no competing interests to declare.
\end{credits}

% \newpage
\bibliographystyle{splncs04}
\bibliography{references}

\end{document}